\documentclass[mnsc,sglanonrev]{informs4}
\RequirePackage{tgtermes}
\RequirePackage{newtxtext}
\RequirePackage{newtxmath}
\RequirePackage{bm}
\RequirePackage{endnotes}

\OneAndAHalfSpacedXI 

\usepackage{algorithm}
\usepackage{array}
\usepackage{centernot}
\usepackage{algpseudocode}
\usepackage{tikz}

\usepackage{natbib}
 \bibpunct[, ]{(}{)}{,}{a}{}{,}%
 \def\bibfont{\small}%
\usepackage{enumitem}
\usepackage{bbm}
\usepackage{multirow}
\usepackage{mathtools}
\usepackage{makecell}
\usepackage{subcaption}
\usepackage{color}
\usepackage{natbib,graphicx,lscape,longtable,epsfig}
\usepackage{amsmath,amssymb}
\usepackage{import}
\usepackage{booktabs}
\usepackage{bm}
\usepackage{url}
\usepackage{etoolbox}
\usepackage{hyperref}
\usepackage{enumitem}
\usepackage{comment}

\def\bs{\boldsymbol}

\def\DM{\mathrm{DM}}

\def\Rate{\mathrm{Rate}}
\def\DP{\mathrm{DP}}
\def\EO{\mathrm{EO}}
\def\PE{\mathrm{PE}}
\def\GSR{\mathrm{GSR}}
\def\TPR{\mathrm{TPR}}
\def\FPR{\mathrm{FPR}}

\usepackage{pdflscape}
\usepackage{tcolorbox}
\usepackage{rotating} 
\tcbuselibrary{breakable}
\usepackage{tabularx}

\newtcolorbox{mybox}{%
    colback=yellow!8,      
    colframe=black,       
    coltitle=gray!10,       
    boxrule=0.1mm,        
    arc=2mm,              
    width=\textwidth,     
    boxsep=1pt,          
    breakable             
}

\newtcolorbox{mybox2}{%
    colback=cyan!3,      
    colframe=black,       
    coltitle=gray!10,       
    boxrule=0.1mm,        
    arc=2mm,              
    width=\textwidth,     
    boxsep=1pt,          
    breakable             
}

\makeatletter
\newcommand{\appendixnumbering}{%

  \renewcommand{\thesection}{\Alph{section}}
  \renewcommand{\theHsection}{appendix.\Alph{section}}

  \renewcommand{\thefigure}{\Alph{section}\arabic{figure}}
  \renewcommand{\thetable}{\Alph{section}\arabic{table}}
  \@addtoreset{figure}{section}
  \@addtoreset{table}{section}
  \renewcommand{\theHfigure}{appendix.\Alph{section}.\arabic{figure}}
  \renewcommand{\theHtable}{appendix.\Alph{section}.\arabic{table}}

  \@addtoreset{theorem}{section}
  \@addtoreset{lemma}{section}
  \@addtoreset{corollary}{section}
  \@addtoreset{remark}{section}
  \@addtoreset{proposition}{section}
  \@addtoreset{definition}{section}

  \renewcommand{\thetheorem}{\Alph{section}\arabic{theorem}}
  \renewcommand{\thelemma}{\Alph{section}\arabic{lemma}}
  \renewcommand{\thecorollary}{\Alph{section}\arabic{corollary}}
  \renewcommand{\theremark}{\Alph{section}\arabic{remark}}
  \renewcommand{\theproposition}{\Alph{section}\arabic{proposition}}
  \renewcommand{\thedefinition}{\Alph{section}\arabic{definition}}
  \renewcommand{\theHtheorem}{appendix.\Alph{section}.\arabic{theorem}}
  \renewcommand{\theHlemma}{appendix.\Alph{section}.\arabic{lemma}}

  \numberwithin{equation}{section}
  \renewcommand{\theequation}{\Alph{section}.\arabic{equation}}
  \renewcommand{\theHequation}{appendix.\Alph{section}.\arabic{equation}}

  \setcounter{figure}{0}
  \setcounter{table}{0}
  \setcounter{theorem}{0}
  \setcounter{lemma}{0}
  \setcounter{corollary}{0}
  \setcounter{remark}{0}
  \setcounter{proposition}{0}
  \setcounter{definition}{0}
  \setcounter{equation}{0}
}
\makeatother

\newcommand{\alignA}{\textsc{Aln}}   

\usepackage{xcolor}

\usepackage{hyperref}
\hypersetup{
	colorlinks=true,
	filecolor=blue,
	citecolor = blue,
	urlcolor=cyan,
}

\usepackage{cases}

\EquationsNumberedThrough    

\TheoremsNumberedThrough     
\ECRepeatTheorems  %

\MANUSCRIPTNO{MNSC-0001-2024.00}

\usepackage[left=1in,right=1in,top=0.9in,bottom=0.9in]{geometry}

\begin{document}


\RUNAUTHOR{Yang}

\RUNTITLE{Fairness May Backfire}

\TITLE{Fairness May Backfire: When Leveling-Down Occurs in Fair Machine Learning}

\ARTICLEAUTHORS{%
\AUTHOR{Yi Yang}
\AFF{Department of Information Systems,
Arizona State University, \EMAIL{yi.yang.10@asu.edu}}

\AUTHOR{Xiangyu Chang}
\AFF{Department of Information Systems and Intelligent Business,
Xi'an Jiaotong University, \EMAIL{xiangyuchang@xjtu.edu.cn}}

\AUTHOR{Pei-yu Chen\thanks{Corresponding author.}}
\AFF{Department of Information Systems,
Arizona State University, \EMAIL{peiyu.chen@asu.edu}}

} 

\ABSTRACT{%
As machine learning (ML) systems increasingly shape access to credit, jobs, and other opportunities, the fairness of algorithmic decisions has become a central concern. 
Yet it remains unclear when enforcing fairness constraints in these systems genuinely improves outcomes for affected groups or instead leads to ``leveling down," making one or both groups worse off. 
We address this question in a unified, population-level (Bayes) framework for binary classification under prevalent group fairness notions. 
Our Bayes approach is distribution-free and algorithm-agnostic, isolating the intrinsic effect of fairness requirements from finite-sample noise and from training and intervention specifics. 
We analyze two deployment regimes for ML classifiers under common legal and  governance constraints: attribute-aware decision-making (sensitive attributes available at decision time) and attribute-blind decision-making (sensitive attributes excluded from prediction).
We show that, in the attribute-aware regime, fair ML necessarily (weakly) improves outcomes for the disadvantaged group and (weakly) worsens outcomes for the advantaged group. 
In contrast, in the attribute-blind regime, the impact of fairness is distribution-dependent: fairness can benefit or harm either group and may shift both groups' outcomes in the same direction, leading to either leveling up or leveling down. 
We characterize the conditions under which these patterns arise and highlight the role of ``masked" candidates in driving them. 
Overall, our results provide structural guidance on when pursuing algorithmic fairness is likely to improve group outcomes and when it risks systemic leveling down, informing fair ML design and deployment choices.
}%





\maketitle


\section{Introduction}

Machine learning (ML) algorithms are increasingly used to make decisions with far-reaching consequences in areas such as credit, healthcare, and criminal justice.
Alongside their growing use, concerns about algorithmic fairness have intensified, particularly when algorithmic outcomes may disadvantage groups defined by sensitive attributes such as race or  gender \citep{barocas2023fairness}.
Addressing fairness in ML has therefore become an important focus of research over the past decade \citep{caton2024},   with presumptive aim to improve outcomes for  disadvantaged or marginalized groups \citep{mittelstadt2023unfairness}.

In response, a large literature has focused on outcome fairness in ML models, especially in classification settings. 
The first line of work has led to several group fairness notions, such as \emph{Demographic Parity} \citep{Dwork2012}, \emph{Equal Opportunity} \citep{Hardt2016}, and \emph{Predictive Equality} \citep{Corbett2017}.
Building on these notions, the second line of work develops various algorithmic interventions, via pre-processing, in-processing, or post-processing approaches, to operationalize fairness \citep{barocas2023fairness}.

\begin{figure}[!ht]
{\centering
\includegraphics[width=0.75\linewidth]{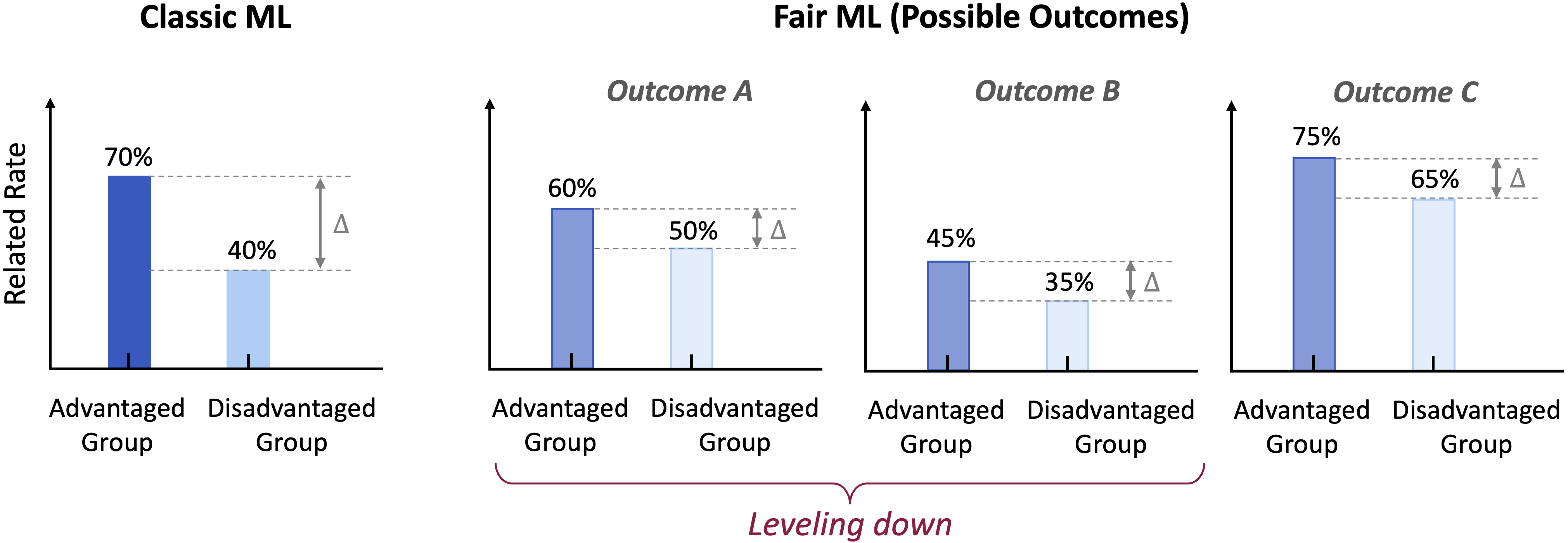}\par}

{\footnotesize\raggedright\noindent \textit{Note.} $\Delta$ denotes the level of (un)fairness under a common fairness notion.\par}

\caption{Outcomes under classical and fairness-aware ML.}
\label{fig:bar_classic_vs_fair}
\end{figure}

Despite these advances in fair ML solutions, recent empirical evidence \citep{mittelstadt2023unfairness,corbett2023measure,jorgensen2023not,yang2024limits} shows that enforcing current fairness notions in practice frequently leads to a \emph{leveling down} phenomenon, where fairness is achieved by making one or both groups worse off (e.g., Outcomes A or B in Figure \ref{fig:bar_classic_vs_fair}).
This raises fundamental questions that remain unresolved: \emph{when are observed fairness gains accompanied by drops in group-specific outcome? When do such decline constitute true cases of leveling down rather than incidental artifacts?  Does leveling down arise systematically from fairness constraints, or only under particular conditions?} \citep{mittelstadt2023unfairness}.

\vspace{-0.5em}
\vspace{-1.1em}

We study these questions in two deployment regimes for ML classifiers that reflect prevailing legal and governance constraints in real-life applications.
In many domains such as credit and employment, laws and regulations against \emph{disparate treatment} prohibit the explicit use of sensitive attributes (e.g., race and gender) in constructing algorithmic predictions (including threshold selection) \citep{kallus2022assessing}.
In some jurisdictions, even collecting or processing such attributes is restricted or prohibited \citep{GDPR2016art}.\footnote{Beyond legal restrictions, the sensitive attribute can be missing at decision time, either because organizations do not collect it or because applicants refuse to disclose it.}
This setting is referred to as \textit{attribute-blind} \citep{chen2024posthoc,yang2025bayes}, in which predictions must depend only on non-sensitive features.
In some other domains (e.g., clinical decision making),
it may be justifiable to incorporate sensitive attribute alongside non-sensitive features in predictions, thus can be permitted by laws \citep{GDPR2016art}. 
For example, population differences in genetic risk or disease pathogenesis may necessitate the inclusion of sensitive attributes (e.g., gender) for prediction  \citep{tannenbaum2019sex}. 
This is  referred to as the \textit{attribute-aware} setting \citep{chen2024posthoc}, in which sensitive attributes can also be used for prediction.
Note that both two regimes can pursue the same fairness notions and develop fair ML models accordingly.\footnote{In the attribute-blind setting,   sensitive attributes are  excluded.
However, this provides only procedural fairness and cannot guarantee outcomes fairness \citep{yang2025toward}. 
Model outcomes can still reflect discrimination along these sensitive attributes due to proxy variables \citep{barocas2016big} or redundant encodings \citep{Hardt2016}. 
Thus, outcome fairness has also been studied in the attribute-blind regime  \citep{Zafar2017fairness2,menon2018cost,chen2024posthoc,yang2025bayes}.
}
However, they implement fairness through different mechanisms. 
As we show in this work, these mechanisms, together with  problem context, govern when, how, and why leveling down occurs.

In this work, we adopt a unified theoretical population-level   framework centered on Bayes-optimal classifiers. 
The Bayes-optimal classifier is the one that attains  the lowest possible average risk (i.e., the theoretical best solution) for a given learning problem: 
thus, no algorithm, no matter how clever or sophisticated, can attain lower average risk than it. 
Bayes-optimal classifiers are foundational in the study of standard binary classification \citep{devroye2013probabilistic}  and serve as a ``limit of what is achievable." 
Since standard learning procedures aim to recover the Bayes-optimal classifier from finite samples, our Bayes lens isolates the intrinsic effects of fairness constraints from finite-sample noise and algorithmic implementation choices (e.g.,  fairness intervention types or training procedure).
Consequently, our results are \emph{structural}, \emph{distribution-free}, and \emph{algorithm-agnostic}--i.e., they hold for arbitrary data-generating processes\footnote{Our results are distribution-free up to a standard regularity condition ensuring well-defined decision thresholds, as is common in statistical machine learning. 
In particular, we do not impose any parametric form on the data distribution.} and do not hinge on any specific training or fairness intervention, thereby revealing the genuine cases of ``leveling down" in fair ML. 

Throughout, we contrast attribute-aware and attribute-blind deployments. 
In each regime, we first characterize the Bayes-optimal (fair) classifier and then trace their effects for different groups.
We then reveal the underlying mechanisms by which fairness is achieved and outcomes are altered via \emph{leveling down}.   
\textbf{Our study yields three main findings and makes several contributions.}
{\textbf{First}, we provide a theoretical, distribution-free, and algorithm-agnostic framework for assessing the effect of fair ML under prevalent fairness notions on the groups affected by these decisions. }
Built on Bayes-optimal rules, our analysis avoids prior distributional assumptions that can shape observed impact patterns and noise from intervention methods, thereby identifying the intrinsic effect of the fairness requirement.
\textbf{Second}, we consider two broader deployment regimes in practice (attribute-aware and attribute-blind) and theoretically show that fair ML operates and affects groups differently across them.
In the attribute-aware regime, fairness redistributes decisions in a way that never harms the disadvantaged group and never benefits the advantaged group.
In the attribute-blind regime, impacts are distribution-dependent: fairness can help or harm either group and may move both groups in the same direction, leading to leveling up or leveling down.
\textbf{Third}, within the attribute-blind  regime,  we further characterize  
the conditions that lead to these patterns and clarify the underlying mechanism,  highlighting the role of \emph{masked candidates}. 
These insights offer  practical guidance for decision-makers on fair ML algorithm design and model alignment.

The remainder of the paper is organized as follows. 
Section~\ref{sec:Related_Work} reviews the related literature. 
Section~\ref{sec:Preliminaries} introduces the problem setup and preliminaries. 
In Section~\ref{sec:FairML_aware}, we analyze  the fair ML impacts in the attribute-aware regime, and Section~\ref{sec:FairML_blind} analyzes the attribute-blind regime.
Proofs and simulations are provided in the full version.

\section{Related Work}\label{sec:Related_Work}

This paper builds on the literature on \emph{Bayes-optimal fair classification}, which characterizes optimal decision rules under fairness constraints \citep{menon2018cost} in both attribute-aware \citep{Corbett2017,schreuder2021classification,chen2024posthoc} and attribute-blind settings \citep{agarwal2018reductions,chen2024posthoc,zeng2024bayes,yang2025bayes}.
Our work adopts these Bayes-optimal characterizations but shift the focus from \emph{algorithmic feasibility} to the \emph{economic consequences} of enforcing fairness:
how decisions are reallocated across groups, how welfare statistics change under fairness  constraints, and when, how, and why \emph{leveling down} can arise even under optimal implementation.

On the economic-impact side, a growing literature studies the downstream consequences of fair ML. 
One line of work provides empirical evidence that enforcing standard fairness notions can yield uneven welfare effects across stakeholders and may even harm the disadvantaged group it aims to protect \citep{jorgensen2023not,mittelstadt2023unfairness}. 
These findings suggest that fairness gains may come with leveling down. 
Motivated by this evidence, we develop a theoretical foundation for these patterns and clarify the underlying mechanisms.

Another line of work analyzes the theoretical impact of fair ML on group outcomes. 
Closer to our setting, several works examine classical binary classification.
{For instance, \citet{liang2051algorithm} and \citet{zhao2022inherent} primarily analyze fairness-accuracy frontiers under Accuracy Parity and Demographic Parity, respectively, and sidely observe that enforcing these two notions raises at least one group's error rate.} 
The most closely related works to ours are \citet{liu2018delayed} and \citet{fu2022Unfair}.
\citet{liu2018delayed} investigate how enforcing strict DP and
EO affects groups' score compared to \textit{equal treatment} (ET) via a single threshold, while \citet{fu2022Unfair} use an analytical model to assess whether groups are better off when a profit-maximizing firm is required to implement strict EO rather than ET, explicitly accounting for the firm's learning effort. 
They adopt a similar score-and-threshold pipeline:
decision-makers use ML algorithms to assigns quality scores to  candidates, assuming the  scores are either uniformly distributed \citep{fu2022Unfair} or satisfy certain independence conditions \citep{liu2018delayed}. 
They then post-process the scores  with a unified threshold (ET) or choosing group-wise thresholds (EO/DP)  to make decisions and compare the resulting group-level impacts.
However, these works adopt a structured, tractable model for the score distribution.
We complement and extend this line of work by allowing more general data-generating environments and by explicitly considering settings in which sensitive attributes are unavailable at decision time (e.g., due to operational or legal constraints), thereby broadening the scope of feasible fairness interventions.
Building on these insights, we provide a Bayes-optimal, population-level characterization of fair-ML impacts that is structural and algorithm-agnostic, and does not rely on a specific parametric model for the underlying data distribution.

\section{Preliminaries}\label{sec:Preliminaries}

\subsection{Classification}

We consider a classification problem specified by a joint distribution $\mathcal D$ over $(X,S,Y)$, where $X\in\mathcal X$ denotes the non-sensitive feature vector, $S\in\mathcal S=\{0,1\}$ a sensitive attribute, and $Y\in\{0,1\}$ the ground-truth label.\footnote{In other words, we focus on binary classification with a single sensitive attribute.
} 
A classifier produces a prediction $\hat Y\in\{0,1\}$ from the available inputs. Depending on the deployment regime, predictions may be based on both $(X, S)$ (attribute-aware) or on $X$ alone (attribute-blind).

We analyze both regimes and investigates how fairness requirements operate differently in each.
To unify notation, let the input space be $\mathcal V$, with $\mathcal V=\mathcal X\times\mathcal S$ in the attribute-aware regime and $\mathcal V=\mathcal X$ in the attribute-blind regime.
We consider a measurable \emph{randomized classifier} $f:\mathcal V\to[0,1]$ that, given $V\in\mathcal V$, outputs a probability of predicting $\hat Y=1$ based on $V$.\footnote{Equivalently, $V=(X,S)$ in the attribute-aware regime and $V=X$ in the attribute-blind regime.}
Let $\mathcal F$ denote the class of such functions and let $\mathrm{Bern}(p)$ denote the Bernoulli distribution with success probability $p\in[0,1]$.

\begin{definition}[Randomized Classifier]\label{df:randomized_classifier} A randomized classifier $f \in \mathcal{F}$ specifies, for any $v \in \mathcal{V}$, the probability $f(v)$ of predicting $\hat{Y} = 1$ given $V = v$, i.e., $\hat{Y} \mid V=v \sim \text{Bern}(f(v))$. \end{definition}

Typically, the quality of a classifier is evaluated using a statistical risk function $R(\cdot; \mathcal{D}) : \mathcal{F} \to \mathbb{R}_{+}$. 
A canonical risk is the cost-sensitive risk \citep{menon2018cost}, defined as follows:

\begin{definition} \label{df:CS_risk}
For a cost parameter $c \in [0, 1]$,
the (expected) cost-sensitive risk is given by  
 $R_{cs}(f;c)=(1-c) P(\hat{Y}=0, Y=1)+c P(\hat{Y}=1, Y=0),$
where $\hat{Y}$ is the prediction produced by $f$, and 
$Y$ is the true label. 
\end{definition}

The cost-sensitive risk allows asymmetric penalization of false negatives and false positives via the cost parameter $c$. 
When $c=0.5$, it reduces to the conventional error rate with equal penalties.

For a given problem, the \textit{Bayes-optimal classifier} is theoretically the best method, achieving the lowest possible average risk.
It is defined as any minimizer $ f^{*} \in \argmin_{f\in\mathcal{F}} R_{cs}(f ; c)$.
In practice, standard learning paradigms aim to recover $f^{*}$ from finite data (e.g., via empirical risk minimization with proper surrogate losses). 
Under mild regularity conditions, such learning algorithms are  {Bayes-consistent}, meaning that the learned classifiers converge to $f^{*}$ as the sample size grows.

Let $\mathbf{1}[\cdot]$ denote the indicator function. 
The following result by \citet{menon2018cost, zeng2024bayes} and \citet{yang2025bayes} characterizes the unconstrained Bayes-optimal classifiers that minimize $R_{cs}(f; c)$.
Its form is given in terms of the posterior class-probability  $\eta(v) \coloneqq P(Y = 1 \mid V=v)$ and  $c$.
\begin{lemma}[Bayes-optimal  Classifiers]\label{lem:bayes_both}
For a cost parameter $c \in [0, 1]$, all Bayes-optimal classifiers $ f^{*}(v) \in \argmin_{f\in\mathcal{F}} R_{cs}(f ; c)$ have the form 
$f^{*}(v)=\mathbf{1}\left[h(v)>0\right]+\alpha\cdot\mathbf{1}\left[h(v)=0\right],$
for all $v\in\mathcal{V}$, where $h(v)=\eta(v)-c$, and $\alpha \in [0, 1]$ is an arbitrary parameter.\footnote{In the following analysis, we set $\alpha=0$ arbitrarily \citep{elkan2001foundations,devroye2013probabilistic,natarajan2018cost}. 
Since for instances at the threshold boundary, the risk is a constant and thus any prediction is optimal.}
 
\end{lemma}


\subsection{Fairness-Aware Classification}
\subsubsection{Fairness Constraints}

Conventional learning objectives minimize expected risk (cost) without accounting for fairness. 
However, extensive empirical evidence documents systematic disparities in algorithmic outcomes across groups defined by sensitive attributes \citep{Obermeyer2019dissecting}. In response, a large literature has formalized algorithmic fairness notions to mitigate these disparities.
For each group $s \in \{0,1\}$, we define Group Selection Rate (GSR): $\GSR_s(f) = {P}(\hat{Y}=1 \mid S=s)$, True Positive Rate (TPR): $\TPR_s(f) = {P}(\hat{Y}=1 \mid Y=1, S=s)$, and False Positive Rate (FPR): $\FPR_s(f) = {P}(\hat{Y}=1 \mid Y=0, S=s)$.\footnote{We consider the non-degenerate case where each group
contains individuals with both labels, that is,
$P(S=s,Y=y) > 0$ for $\forall s\in\{0,1\}$ and $\forall y\in\{0,1\}$. 
Then the group-wise TPR and FPR are well-defined.} 
Then, we consider the following fairness notions widely used in ML.

\begin{definition}[Demographic Parity]  
A classifier $f$ satisfies \textit{demographic parity}  \citep{Dwork2012} if its prediction $\hat{Y}$ is  independent of the sensitive attribute $S$, so that  
$\GSR_0(f) = \GSR_1(f)$.
\end{definition}

\begin{definition}[Equal Opportunity]  
A classifier $f$ satisfies \textit{equal opportunity} \citep{Hardt2016} if it achieves the same true positive rate between groups, i.e., 
$\TPR_0(f) = \TPR_1(f)$. 
\end{definition}

\begin{definition}[Predictive Equality]  
A classifier $f$ satisfies \textit{predictive equality} \citep{Corbett2017} if it achieves the same false positive rate between groups, i.e., $\FPR_0(f) = \FPR_1(f)$.
\end{definition}

In practice, achieving the equalities above (i.e., perfect fairness) often requires a significant sacrifice in efficiency (higher expected risk) and is typically infeasible with finite data \citep{Shen2023}. 
Instead, a limited level of disparity may be preferred. 
Accordingly, previous research has typically measures fairness as differences in the quantities that would be equal under perfect fairness and focuses on minimizing risk subject to constraints on disparity levels  \citep{Zafar2019fairness,williamson2019fairness, chen2024posthoc, denis2024fairness}. 
Following this line of work, we define disparity measures as group differences in the quantities equalized by each fairness notion.

\begin{definition}[Disparity Measures]\label{def:Disparity_Measures}
For a randomized classifier $f$ with prediction $\hat Y$, let $y^\star\subseteq\{0,1\}$.
For $s\in\{0,1\}$, let
$\Rate_s(f) =P(\hat Y=1 \mid S=s,  Y\in y^\star)$ denote the notion-conditioned selection rate for group $s$ under $f$. 
Then, the  \textit{disparity measure} for $f$ is defined as:  
$\DM(f) = \Rate_1(f)-\Rate_0(f).$  
The value of $y^\star$ depends on the chosen fairness notion: For DP, $y^\star=\{0,1\}$; for EO, $y^\star=\{1\}$; and for PE, $y^\star=\{0\}$.
\end{definition}

We refer to the notion-conditioned selection rate for group $s$, $\Rate_s(f)$, as the \emph{notion-target rate} (NTR), since it is the statistic that the chosen fairness notion targets for equalization across groups. 
Intuitively, $\DM(f)$ measures the signed gap in NTRs between two groups; its sign indicates which group attains the higher NTR under the chosen notion (and thus, which group benefits more) when applying a classifier $f$.

\begin{remark}\label{rmk:group_adv_disadv}
Usually, the event $\hat{Y}=1$ is considered a favorable action,\footnote{This can always be achieved by encoding the favorable outcome in the ground-truth label as $Y=1$. 
} 
since an individual is selected (e.g., granting a loan \citep{fu2020artificial}). 
In this case, for a chosen notion, the \textit{advantaged group} is the group with a higher NTR, since the model favors them relative to the other.
We refer to the other group as the \emph{disadvantaged group}.
\end{remark}

\subsubsection{Byes-Optimal Fair Classifier}
In fairness-aware classification, decision-makers typically permit a limited amount of disparity and pre-specify an unfairness tolerance level $\delta\in[0,1]$.\footnote{Note that $\delta=0$ corresponds to perfect fairness; larger $\delta$ relaxes the constraint and indicates less fairness.}
For a given $\delta$, they seek a \emph{Bayes-optimal fair} classifier $f^{*}_\DM(v)$ that minimizes the cost subject to its unfairness level not exceeding $\delta$.
Formally, this can be expressed as:
\begin{equation}\label{eq:bayes_fair_prob}
f^{*}_\DM\in\argmin_{f\in\mathcal{F}} \left\{
R_{cs}(f ; c): |\DM(f)| \leq \delta
\right\}.    
\end{equation}
Prior work \citep{menon2018cost, zeng2024bayes, yang2025bayes} also characterizes the form of Bayes-optimal fair classifiers, as presented in Lemma~\ref{lem:fair_bayes_both}.

\begin{lemma}[Bayes-optimal Fair Classifiers]\label{lem:fair_bayes_both}
Let $c\in[0,1]$, $\delta\in[0,1]$, and a fairness notion $\DM\in\{\DP,\EO,\PE\}$.
Let $\mathcal V$ denote the input space. 
Then, there exists $\lambda^{*}\in\mathbb R$\footnote{The value of $\lambda^{*}$ corresponds to the difference of Lagrange multipliers for the two bounds in $|\DM(f)|\le\delta$ and may be negative \citep{menon2018cost}. 
Its exact value depends on the data distribution $\mathcal D$, the chosen fairness notion, and $\delta$. 
In other words, for the constrained problem $\min\{R_{cs}(f;c):|\DM(f)|\le\delta\}$, there always exists $\lambda^{*}\in\mathbb R$ such that $f^{*}_\DM(v)$ attains the minimum.} such that all Bayes-optimal fair classifiers $f^{*}_\DM(v)\in\mathcal F$ that minimize $\{R_{cs}(f;c):|\DM(f)|\le\delta\}$ have the form of 
$f^{*}_\DM(v)
=
\mathbf 1\left[h_\DM(v)>0\right]
+
\alpha\cdot \mathbf 1\left[h_\DM(v)=0\right],$
for all $v\in\mathcal V$.
Here, 
$h_\DM(v) =\eta(v)-c-\lambda^{*}\cdot\nu_\DM(v),$ $\eta(v) = P(Y = 1 \mid V=v)$, and $\alpha \in [0,1]$ is arbitrary.
The term $\nu_\DM(v)$ depends on the chosen fairness notion.
In the attribute-aware regime, $\mathcal V=\mathcal X\times\mathcal S$, and in the attribute-blind regime, $\mathcal V=\mathcal X$.
\end{lemma}

Lemmas~\ref{lem:bayes_both} and~\ref{lem:fair_bayes_both}  show that, in either regime, the Bayes-optimal fair rule compares the posterior $\eta$ to the unconstrained threshold $c$ adjusted by a fairness correction $\nu_\DM$. 
Note that in this paper, we consider the setting where the random variables $\eta(X)$, $\nu_{\DM}(X)$, 
$\eta(X,s)$ and $\nu_{\DM}(X,s)$ for $\forall s\in\mathcal S$, admit probability
density functions (i.e., have continuous distributions) over $\mathcal{X}$ \citep{ lei2014classification,chzhen2019leveraging,zeng2024bayes}.
This implies  the boundary sets of randomized classifiers are of measure zero \citep{chzhen2019leveraging}.

\section{Fair ML Impacts in the Attribute-Aware Regime}\label{sec:FairML_aware}

In the attribute-aware regime, with $\mathcal{V}=\mathcal{X}\times\mathcal{S}$, Lemma~\ref{lem:bayes_both} gives the unconstrained Bayes-optimal classifier:
\begin{equation}\label{eq:simplified_bayes_f_S}
f^{*}(x,s)=\mathbf{1}\left[\eta(x,s)>c\right].
\end{equation} 
By \citet{menon2018cost,zeng2024bayes,yang2025bayes} and Lemma~\ref{lem:fair_bayes_both}, the Bayes-optimal fair classifier enforcing $\DM\in\{\DP,\EO,\PE\}$ at tolerance $\delta\in[0,1]$ is:
\begin{equation}\label{eq:simplified_bayes_fairf_S}
f_\DM^{\lambda^*}(x, s)=\mathbf{1}\left[\eta(x,s)>c+\lambda^{*}\cdot\nu_\DM(x,s) \right],    
\end{equation}
where 
$\nu_\DP(x,s) = \frac{(2s - 1)}{P(S = s)}$, $\nu_\EO(x,s) = \frac{(2s - 1)\eta(x,s)}{P(S = s, Y = 1)},$ and $\nu_\PE(x,s) = \frac{(2s - 1)(1-\eta(x,s))}{P(S = s, Y = 0)}. $
Note that if $\lambda^{*}=0$, then $f_\DM^{\lambda^*}(x,s)=f^{*}(x,s)$; that is, the unconstrained classifier already satisfies the fairness constraint. 
Moreover,  let $s_h:= \argmax_{s \in \mathcal{S}} \Rate_s(f^*(x,s))$ denote the advantaged group and $s_l: = \argmin_{s \in \mathcal{S}} \Rate_s(f^*(x,s))$ the disadvantaged group (Remark~\ref{rmk:group_adv_disadv}).  
We now analyze how enforcing fairness affects these two groups.

\subsection{Impact on Different Groups}\label{sec:impact_two_reoups_S}

Enforcing fairness in the aware regime introduces a correction term $\nu_\DM(x,s)$ that yields group-specific thresholding: individuals with the same $x$ can be treated differently across groups.
For all three  notions, $\nu_\DM(x,1)\ge 0$ and $\nu_\DM(x,0)\le 0$ for all $x$, so the fair threshold $c+\lambda^{*}\nu_\DM(x,s)$ shifts in opposite directions across groups relative to the unconstrained threshold $c$. 
This is formalized in Proposition~\ref{pp:threshold_shifts_S}.
\begin{proposition}[Opposite  Threshold Shifts Across Groups]
\label{pp:threshold_shifts_S}
For a  notion $\DM \in \{\DP, \EO, \PE\}$ and a tolerance level $\delta\in[0,1]$, let $\lambda^{*}$ be its corresponding optimal multiplier in \eqref{eq:simplified_bayes_fairf_S}.
Then, we have, for all $x\in\mathcal X$,
$c+\lambda^{*}\nu_\DM(x,s_h)\ \ge\ c$  and $
c+\lambda^{*}\nu_\DM(x,s_l)\ \le\ c,$
with equality for both groups \textit{iff}  $\lambda^{*}=0$.
\end{proposition}

Proposition~\ref{pp:threshold_shifts_S} shows that, in the attribute-aware regime, fairness weakly increases the threshold for the advantaged group and weakly decreases it for the disadvantaged group relative to $c$, with equality \emph{iff} the unconstrained classifier already satisfies the fairness constraint. Thus, fair ML systematically shifts decisions toward the disadvantaged group: the selected region (weakly) contracts for $s_h$ and expands for $s_l$.

We then examine how fairness requirements affect group-level outcomes by tracking each group's NTR, $\Rate_s(f)$, as we move from $f^{*}(x,s)$ to  $f_\DM^{\lambda^*}(x, s)$.
Theorem~\ref{thm:aware_work_as_want} characterizes the resulting NTR shifts.

\begin{theorem}[NTR Redistribution with $S$]\label{thm:aware_work_as_want}
Fix a fairness notion $\DM \in \{\DP, \EO, \PE\}$ and a tolerance level $\delta\in[0,1]$. 
Let $\Rate_{s}$ be the group-wise NTR under $\DM$. 
With $f^{*}$ as in \eqref{eq:simplified_bayes_f_S} and $f_\DM^{\lambda^{*}}$ as in \eqref{eq:simplified_bayes_fairf_S}, we have
$\Rate_{s_{h}}(f_\DM^{\lambda^{*}})\leq\Rate_{s_{h}}(f ^{*})  $   and $\Rate_{s_{l}}(f_\DM^{\lambda^{*}})\geq\Rate_{s_{l}}(f ^{*})$.
\end{theorem}

Theorem~\ref{thm:aware_work_as_want} shows that, in the attribute-aware regime, fair classifiers shift group NTRs in opposite directions toward equalization: they weakly improve outcomes for the disadvantaged group while potentially weakly reducing outcomes for the advantaged group. 
This pattern always holds and aligns with the presumptive aim of the field: to improve outcomes for disadvantaged groups that unjustifiably receive worse treatment or outcomes than their peers \citep{mittelstadt2023unfairness}.

Having established how each group's NTR responds at a fixed tolerance level $\delta$, we next examine how the magnitude of this redistribution varies with $\delta$.
Knowing this effect helps calibrate $\delta$ and provides an overview of how it shapes group-level outcomes.
Proposition~\ref{pp:delta_amplify_S} provides the comparative statics.

\begin{proposition}[Lower $\delta$ Always  Amplifies Aware Fair ML's Impact] \label{pp:delta_amplify_S}
Under the same setup as in Theorem~\ref{thm:aware_work_as_want}, as a function of $\delta$  (with $\lambda^{*}=\lambda^{*}(\delta)$), we have that
$\Rate_{s_{h}}(f_\DM^{\lambda^{*}})$ is monotone non-decreasing, and 
$\Rate_{s_{l}}(f_\DM^{\lambda^{*}})$ is monotone non-increasing.
\end{proposition}

Proposition~\ref{pp:delta_amplify_S} shows that tighter fairness constraints (smaller $\delta$)  
intensifies the opposite-direction adjustments of NTRs across groups.
Specifically, lowering $\delta$ weakly reduces the advantaged group's NTR and weakly raises the disadvantaged group's NTR. 
The effect is always order-preserving in $\delta$.

We next consider group precision, defined for group $s$ as $\pi_{s}(f)=P(Y=1\mid \hat{Y}=1, S=s)$, 
i.e., the fraction of accepted candidates in group $s$ who are truly qualified.
It complements selection rates by capturing the quality (rather than the quantity) of acceptances.
Theorem~\ref{thm:group_precision_S} shows the results.
\begin{theorem}[Group-wise Precision by Fair ML  with $S$]\label{thm:group_precision_S}
For a classifier $f$, let 
$\pi_{s}(f)$ denote the precision of group $s\in\mathcal{S}$ when $P(\hat{Y}=1\mid S=s)>0$. 
Then, for any group $s$ with both $\pi_{s}(f ^{*})$ and $\pi_{s}(f_\DM^{\lambda^{*}})$  well-defined, the following holds: 
$\pi_s(f_\DM^{\lambda^{*}})\ge \pi_s(f^{*})$ if $s=s_h$, and 
$\pi_s(f_\DM^{\lambda^{*}})\le \pi_s(f^{*})$ if $s=s_l$.
\end{theorem}

Theorem~\ref{thm:group_precision_S} implies that  applying the fair classifier always weakly increases precision for the advantaged group and weakly decreases it for the disadvantaged group (when precision is well-defined under both rules). 
Thus, even though the disadvantaged group's selection-based rates rise, the quality of its acceptances declines. 
This is concerning: approving unqualified individuals can harm them (e.g., default and credit-score damage in lending) and, at the group level, can increase observed adverse outcomes, making the disadvantaged group appear riskier \citep{fu2022Unfair}.
Conversely, for the advantaged group, selection-based rates weakly decrease while precision weakly increases, implying higher-quality acceptances and fewer observed bad outcomes. 
This improvement in apparent risk profile comes at the cost of reduced access: some qualified advantaged candidates may be screened out, raising concerns about ``reverse discrimination" \citep{yang2025toward}. 
These directional effects are notion-agnostic in the aware regime and  hold for all $\delta$.

\subsection{Redistribution Mechanism: The Role of the Sensitive Attribute}\label{sec:Aware_Mechanism}
\begin{figure}[!ht]
\centering
\includegraphics[width=0.55\textwidth]{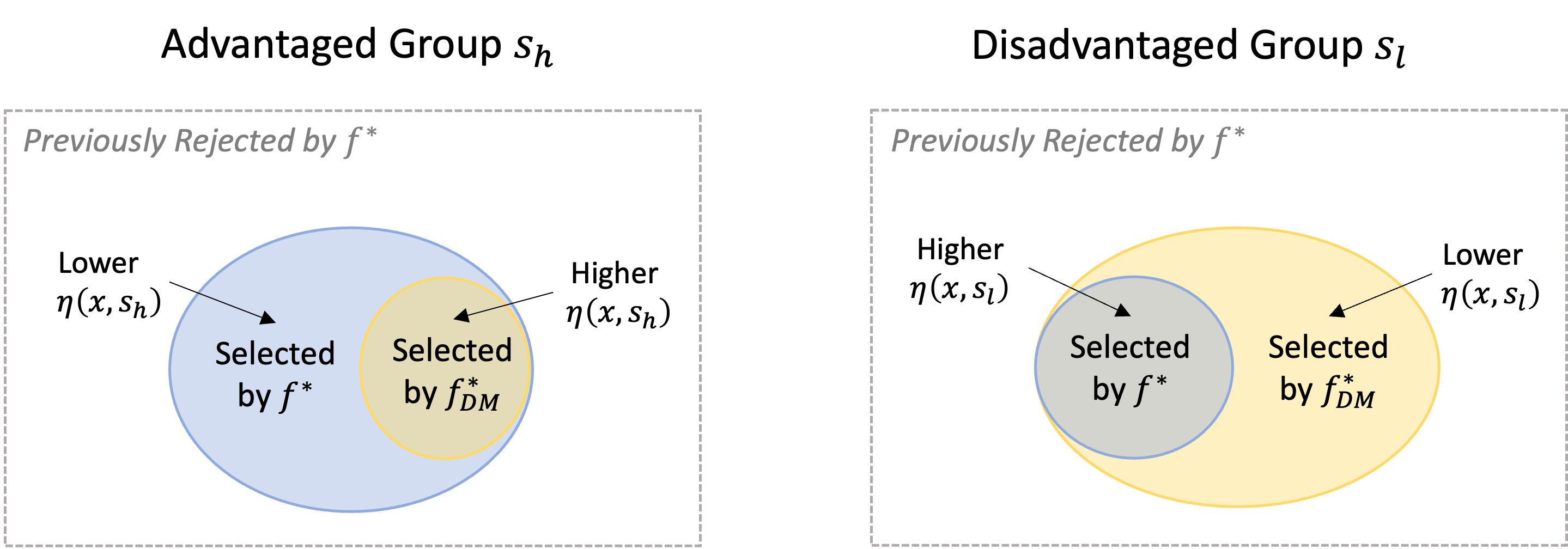}
\caption{Selection Regions Before and After Enforcing Fairness in the Attribute-Aware Regime.}
\label{fig:venn_aware}
\end{figure}

We show that, in the attribute-aware regime, fair ML always has a deterministic direction of impact across the two groups.
These effects are driven by a simple threshold-shift mechanism based on $S$. 
By Proposition~\ref{pp:threshold_shifts_S}, using $S$ in prediction allows group-specific thresholds relative to the unified unconstrained threshold: 
the threshold weakly increases for the advantaged group and weakly decreases for the disadvantaged group. 
Figure~\ref{fig:venn_aware} then illustrates this redistribution.
\paragraph{Advantaged group (deletion only).}
Raising the threshold for group $s_h$ from $c$ removes previously selected candidates in \emph{increasing} order of $\eta(x,s_h)$. 
Hence, all selection rates (including NTR) weakly decrease, while precision weakly increases, since the removed candidates are those less likely to be qualified among the previously selected.

\paragraph{Disadvantaged group (inclusion only).}
Lowering the threshold for group $s_l$ from $c$ adds previously unselected instances in \emph{decreasing} order of $\eta(x,s_l)$.
Hence, all selection rates (including NTR) weakly increase, while precision weakly decreases, as the added candidates are less likely to be qualified than those previously selected.
\vspace{-1.8em}
\vspace{-1.5em}

\section{Fair ML Impacts in the Attribute-Blind Regime}\label{sec:FairML_blind}

In the previous section, we studied the \emph{attribute-aware} setting, where the sensitive attribute is available and can be used in prediction. 
While permissible in some domains (e.g., medical diagnosis), it may be restricted or prohibited in others; for instance, the Equal Credit Opportunity Act bars lending decisions that explicitly depend on race or gender. 
In such settings, models must operate in an \emph{attribute-blind} regime, predicting using only non-sensitive features $X$, so the input space is $\mathcal{V}=\mathcal{X}$. 
In this case, Lemma~\ref{lem:bayes_both} characterizes the unconstrained Bayes-optimal classifier for all $x\in\mathcal{X}$:
\begin{equation}\label{eq:simplified_bayes_f_noS}
f^{*}(x)=\mathbf{1}\left[\eta(x)>c\right].
\end{equation}  
By \citet{menon2018cost,zeng2024bayes,yang2025bayes} and Lemma~\ref{lem:fair_bayes_both}, the Bayes-optimal fair classifier enforcing $\DM\in\{\DP,\EO,\PE\}$ at tolerance $\delta\in[0,1]$ in the blind regime is:
\begin{equation}\label{eq:simplified_bayes_fairf_noS}
f_\DM^{\lambda^*}(x)=\mathbf{1}\left[\eta(x)>c+\lambda^{*}\cdot\nu_\DM(x) \right],    
\end{equation}
where $\nu_\DP(x)  =  \frac{P(S=1 \mid X=x)}{P(S=1)} - \frac{P(S=0 \mid X=x)}{P(S=0)}$, $\nu_\EO(x) = \frac{P(S=1, Y=1 \mid X=x)}{P(S=1, Y=1)} - \frac{P(S=0, Y=1 \mid X=x)}{P(S=0, Y=1)}$, and $\nu_\PE(x) = \frac{P(S=1, Y=0 \mid X=x)}{P(S=1, Y=0)} - \frac{P(S=0, Y=0 \mid X=x)}{P(S=0, Y=0)}.$
Note that if $\lambda^{*}=0$, then $f_\DM^{\lambda^*}(x)=f^{*}(x)$; 
that is, the unconstrained classifier already satisfies the fairness constraint. 
Furthermore,  let $s_h:= \argmax_{s \in \mathcal{S}} \Rate_s(f^*(x))$ denote the advantaged group and $s_l: = \argmin_{s \in \mathcal{S}} \Rate_s(f^*(x))$ the disadvantaged group (Remark~\ref{rmk:group_adv_disadv}) in the blind regime.\footnote{Here, we abuse notation and reuse $s_h$ and $s_l$ for the advantaged and disadvantaged groups in the attribute-blind regime. This does not imply that the same group is advantaged/disadvantaged in both regimes; the identity of the advantaged and disadvantaged groups can differ between the attribute-aware and attribute-blind deployments (i.e., $\argmax_{s \in \mathcal{S}} \Rate_s(f^*(x))=\argmin_{s \in \mathcal{S}} \Rate_s(f^*(x,s))$).}  
We now analyze how enforcing fairness affects these two groups in the blind regime.




\subsection{Impact on Different Groups}

In the attribute-blind regime, the correction term $\nu_\DM(x)$ depends only on $x$ and is $S$-invariant, as shown in \eqref{eq:simplified_bayes_fairf_noS}. 
Thus, individuals with the same $x$ receive the same decision regardless of group membership. 
This contrasts with the attribute-aware regime, where the correction $\nu_\DM(x,s)$ induces group-specific thresholds and hence group-dependent treatment for the same $x$.

Furthermore, unlike the aware regime in which the direction (sign) of the threshold shift is fixed only by $s$ and is distribution-agnostic,
in the blind case the direction of the threshold adjustment at $x$ is governed by the group composition conditional on $X$, as captured by $P(S\mid X=x)$ or $P(S,Y\mid X=x)$.
Consequently, the fairness correction aggregates group information at the feature level rather than at the group label: 
its effect varies with $x$, and group-level outcomes depend on how each group is distributed over $X$. 
Hence, in the attribute-blind regime, group impacts are distribution-dependent rather than predetermined by group membership as in the aware regime.
Below, we first partition candidates into three non-overlapping regions (Definition \ref{def:dis/adv_like_noS}), and then formalize the resulting feature-level threshold shifts in Proposition \ref{pp:threshold_shifts_noS}.

\begin{definition}[Advantaged-like and Disadvantaged-like]\label{def:dis/adv_like_noS}
For a notion $\DM \in \{\DP, \EO, \PE\}$ and a tolerance level $\delta\in[0,1]$,  let $\lambda^{*}$ be the optimal multiplier in \eqref{eq:simplified_bayes_fairf_noS}. 
Define
$\mathcal Q^{h}=\{x\in\mathcal X:\ \lambda^{*}\nu_{\DM}(x)>0\},
\mathcal Q^{l}=\{x\in\mathcal X:\ \lambda^{*}\nu_{\DM}(x)<0\}$, and $\mathcal Q^{\circ}=\{x\in\mathcal X:\ \lambda^{*}\nu_{\DM}(x)=0\}.$
We call points in $\mathcal Q^{h}$ \emph{advantaged-like}, those in $\mathcal Q^{l}$ \emph{disadvantaged-like}, 
and those in $\mathcal Q^{\circ}$ \emph{neutral} 
(in the sense of $\nu_{\DM}(x)$).  
\end{definition}

\begin{proposition}[Feature-level Threshold Shift]\label{pp:threshold_shifts_noS}
Fix a fairness notion $\DM\in\{\DP,\EO,\PE\}$ and a tolerance $\delta\in[0,1]$, and let $\lambda^{*}$ be the optimal multiplier in \eqref{eq:simplified_bayes_fairf_noS}. 
With $\mathcal Q^{h},\mathcal Q^{l},\mathcal Q^{\circ}$ as in Definition \ref{def:dis/adv_like_noS},
for any $x\in\mathcal X$, we have: $c+\lambda^{*}\nu_\DM(x)>c$ if $x\in\mathcal Q^{h}$; $ c+\lambda^{*}\nu_\DM(x)<c$ if $x\in\mathcal Q^{l}$; and $c+\lambda^{*}\nu_\DM(x)=c$ if $x\in\mathcal Q^{\circ}.$
\end{proposition}

Proposition \ref{pp:threshold_shifts_noS} shows that fairness introduces a $x$-dependent correction which increases the threshold (penalizes) for candidates from advantaged-like side and decreases it (subsidizes) those from the disadvantaged-like side.
For candidates whole looks neutral, the threshold is not adjusted.
Hence, the selection set
{shrinks} (relative to that of the unconstrained classifier) on $\mathcal{Q}^{h}$,   {expands} on $\mathcal{Q}^{l}$, and unchanged on $\mathcal{Q}^{\circ}$.
Consequently, the net effect on each group depends on how $P(S\mid X)$ or $P(S,Y\mid X)$ distributes mass over these  regions.

Next, we investigate the NTRs associated with a chosen  notion. 
As noted in Figure \ref{fig:bar_classic_vs_fair}, they can vary via one of three patterns when reducing unfairness: (i) Both groups' NTRs weakly decrease, with a larger drop for the advantaged group;
(ii) Both groups' NTRs weakly increase, with a larger gain for the disadvantaged group;
or (iii) The advantaged group's NTR weakly decreases, while the disadvantaged's weakly increases.\footnote{These patterns may exhibit equalities (i.e., no change) on both groups  only when the unconstrained classifier has already satisfied the fairness constraint (i.e., when $\lambda^*=0$).} 
We will show that all three patterns can arise in the attribute-blind regime and identify  distributional conditions under which each occurs. 
We discuss the first two patterns here and leave the third one in the full version.

\begin{theorem}[NTR Redistribution Without $S$]\label{thm:NTR_redistribution_noS}
Fix a notion $\DM\in\{\DP,\EO,\PE\}$ and $\delta\in[0,1]$.
For $x\in\mathcal{X}$ with $\nu_{\DM}(x)\ne 0$, define
$\zeta(x) = \frac{\eta(x) - c}{\nu_{\DM}(x)}$ 
and the extrema
$A_{\max} = \sup_{\bs{x}   :   \nu_{\DM}(x) > 0} \zeta(x), 
A_{\min} = \inf_{\bs{x}   :   \nu_{\DM}(x) > 0} \zeta(x),
B_{\max} = \sup_{\bs{x}: \nu_{\DM}(x) < 0} \zeta(x), 
B_{\min} = \inf_{\bs{x}:   \nu_{\DM}(x) < 0} \zeta(x).$
For group $s\in\mathcal{S}$, let $\Rate_{s}$ denote its group-wise NTR under $\DM$. 
With $f^{*}$ as in \eqref{eq:simplified_bayes_f_noS} and $f_\DM^{\lambda^{*}}$ as in \eqref{eq:simplified_bayes_fairf_noS}, the following hold for any group $s$:
If $A_{\max} \leq B_{\min}$, 
    then  
    $\Rate_{s}(f_\DM^{\lambda^{*}}) \leq \Rate_{s}(f^{*});$
if $B_{\max} < A_{\min}$, then 
    $\Rate_{s}(f_\DM^{\lambda^{*}}) \geq \Rate_{s}(f^{*}).$
\end{theorem}

Theorem~\ref{thm:NTR_redistribution_noS} shows that in the  blind regime, fair ML can move both groups' NTRs in the same direction.  
Because the blind rule is $S$-invariant, the same set of $x$ flips together; 
how each group's NTR changes depends on the mass each group places on that flipped set.
If that set carries ``similar" share in two groups, or if all flipped points are more characteristic of the same group, i.e., all lie in $\mathcal{Q}^{h}$   ($A_{\max}\le B_{\min}$) or all lie in $\mathcal{Q}^{l}$ ($B_{\max}<A_{\min}$),
both groups' NTRs move in the same direction. 
This contrasts with the aware regime (Theorem~\ref{thm:aware_work_as_want}), where fairness always shifts the two groups in opposite directions toward the target (within tolerance $\delta$). 
In the attribute-blind regime, such opposite-direction equalization need not occur, and the group effects are distribution-dependent.

Theorem~\ref{thm:NTR_redistribution_noS} characterizes group NTRs at a fixed tolerance in the attribute-blind regime, we then study how they respond as the tolerance $ \delta $ varies. 
\begin{proposition}[When Lower $\delta$  Amplifies Blind Fair ML's Impact] \label{pp:delta_amplify_noS}
Under the same setup as in Theorem~\ref{thm:NTR_redistribution_noS}, let $\lambda^* = \lambda^*(\delta)$. 
For any group $s \in \mathcal{S}$:
 If { $A_{\max} \leq B_{\min}$}, then    
    $\Rate_{s}(f_\DM^{\lambda^{*}})$ is monotone non-decreasing in $\delta$; if { $B_{\max} < A_{\min}$}, then 
  $\Rate_{s}(f_\DM^{\lambda^{*}})$ is monotone non-increasing in $\delta$. 
\end{proposition}

Relative to Proposition~\ref{pp:delta_amplify_S} (attribute-aware), where tightening $\delta$ produces opposite-direction, order-preserving shifts across groups for all distributions, Proposition~\ref{pp:delta_amplify_noS} shows that in the attribute-blind regime the response is distribution-dependent. 
Under the separation $A_{\max}\leq B_{\min}$ (resp.\ $B_{\max}<A_{\min}$), tightening $\delta$ weakly lowers (resp.\ raises) both groups' NTRs, and the response is order-preserving in $\delta$. 
When neither separation holds, a uniform direction is not guaranteed, and responses may switch direction as $\delta$ varies. 
For example, there may exist $0\le \delta_{1}<\delta_{2}<|D_{\DM}(0)|$ and a group $s$ such that $\Rate_{s}(f_{\DM}^{\lambda^{*}(\delta_{2})})<\Rate_{s}(f^{*})<\Rate_{s}(f_{\DM}^{\lambda^{*}(\delta_{1})})$. 
Thus, unlike the attribute-aware case, tightening $\delta$ does not further raise the disadvantaged group's NTR (nor lower the advantaged group's NTR) for all distributions; rather, how the gap narrows under tighter fairness is distribution-dependent.

We next analyze how blind fair ML affects group-wise precision, building on the order-alignment conditions in Definition~\ref{def:order_align_con}. 
Theorem~\ref{thm:group_precision_noS} then presents the result.
\begin{definition}[Order Alignment]\label{def:order_align_con}
Let $\mathcal G\subseteq \mathcal{X}$ and 
$g,h:\mathcal{X} \to\mathbb R$ be real-valued functions.  
We say that $g$ is \emph{aligned} with $h$ on $\mathcal{G}$, written
      $g\overset{\alignA}{\sim}h$ on $\mathcal{G}$,
      if $\forall x,x' \in \mathcal G$, $g(x)\leq g(x') \Rightarrow h(x)\leq h(x').$
      
\end{definition}

\begin{theorem}[Group-wise Precision by Fair 
ML Without $S$]\label{thm:group_precision_noS}
For a group $s\in\mathcal{S}$ and a classifier $f$, define the group-wise precision achieved by $f$ as
$\pi_{s}(f)=P(Y=1|\hat{Y}=1, S=s)$  whenever $P(\hat{Y}=1\mid S=s)>0$.
Let $f^{*}(x)$ as in \eqref{eq:simplified_bayes_f_noS} and $f^{\lambda^*}_{\DM}(x)$ as in \eqref{eq:simplified_bayes_fairf_noS}, and let $\zeta(x)$, $A_{\max}, A_{\min}, B_{\max}$ and $B_{\min}$ be as in Theorem~\ref{thm:NTR_redistribution_noS}.
Write $\eta_s(x):=\eta(x,s)$.  
Then, for any group $s$ with both $\pi_{s}(f ^{*})$ and $\pi_{s}(f_\DM^{\lambda^{*}})$ well-defined, the following holds: 
\begin{numcases}{\text{If $A_{\max}\le B_{\min}$, then }\pi_s\!\bigl(f^{\lambda^*}_{\DM}\bigr)\,}
\displaystyle \geq \pi_{s}(f^{*})
  & \text{if } $\zeta \overset{\alignA}{\sim} (2s_h-1)\,\eta_s 
    \text{ on } \mathcal Q^{h}$,
\label{eq:precision_con11}\\
\displaystyle \leq \pi_{s}(f^{*})
  & $\text{if } \zeta \overset{\alignA}{\sim} (1-2s_h)\,\eta_s 
    \text{ on } \mathcal Q^{h}.$ {\color{white}xxxxxxxxxxxxxxxxxxxxxxx}
\label{eq:precision_con12}
\end{numcases}
\begin{numcases}{\text{If $B_{\max}<A_{\min}$, then }\pi_s\!\bigl(f^{\lambda^*}_{\DM}\bigr)\,}
\displaystyle \ge \pi_s(f^{*}), 
  & $\text{if } \sup_{\mathcal Q^{h}}\eta_s
    \le \inf_{\mathcal Q^{l}}\eta_s
    \ \text{and}\
    \zeta \overset{\alignA}{\sim} (2s_h-1)\,\eta_s
    \text{ on } \mathcal Q^{l}$,\label{eq:precision_con21}\\ 
\displaystyle \le \pi_s(f^{*}), 
  & $\text{if } \sup_{\mathcal Q^{l}}\eta_s
    \le \inf_{\mathcal Q^{h}}\eta_s
    \ \text{and}\
    \zeta \overset{\alignA}{\sim} (1-2s_h)\,\eta_s
    \text{ on } \mathcal Q^{l}.$
    \label{eq:precision_con22}
\end{numcases}

\end{theorem}


When the separation conditions hold and precision is well-defined, Theorem~\ref{thm:group_precision_noS} characterizes group-wise precision shifts under blind fair ML. 
If a distribution satisfies $A_{\max}\le B_{\min}$ and the condition in \eqref{eq:precision_con11} for a group $s$, then the flipped-out points have lower $\eta$ than the remaining selected set, so $\pi_s$ weakly increases; 
under \eqref{eq:precision_con12} the flipped-out points have higher $\eta$, so $\pi_s$ weakly decreases. 
If a distribution satisfies $B_{\max}< A_{\min}$ and the condition in \eqref{eq:precision_con21} for a group $s$, then the flipped-in points have higher $\eta$ than the previously selected set, so $\pi_s$ weakly increases;
under \eqref{eq:precision_con22} the flipped-in points have lower $\eta$, so $\pi_s$ weakly decreases.
This contrasts with the attribute-aware regime (Theorem~\ref{thm:group_precision_S}), where precision always moves in opposite directions across groups (advantaged weakly up, disadvantaged weakly down); 
In the blind regime, either direction can occur for each group, and both groups may move the same way (under separation).

\subsubsection{Redistribution Mechanism: The Role of Masked Candidates}\label{sec:Blind_Mechanism}

Recall that in the attribute-blind regime, enforcing fairness introduces an $x$-dependent correction which penalizes individuals on the \emph{advantaged-like} side $\mathcal Q^{h}$ and subsidizes those on the \emph{disadvantaged-like} side $\mathcal Q^{l}$, while leaving the neutral side $\mathcal Q^{\circ}$ unchanged.
Because $S$ is unobserved, each side contains a \emph{mixture} of true-$S=1$ and true-$S=0$ individuals.
In other words, on each side there are candidates who, 
look advantaged-like (resp.\ disadvantaged-like) but whose true group label is the opposite\footnote{A candidate may look advantaged-like by $\nu_{\DM}(x)$ while actually belonging to the disadvantaged group, and vice versa.
Take DP as an example, for a instance $x$, even its $P(S=1|X=x)>P(S=0|X=x)$ (i.e., are Group 1-like), $x$ can actually be long to Group 0.}--these are the \emph{masked candidates}.
Note that in the blind regime, when moving from unfair classifier to the fair one, deletions on the advantaged-like side and additions on the disadvantaged-like side can occur simultaneously on both groups, with net effects set by the data distribution. This contrasts with the aware case, where flips are only group-specific (deletion only on the advantaged and addition only on the disadvantaged). 
In the blind case, if the flipped set is concentrated in one side ($\mathcal Q^{h}$ or $\mathcal Q^{l}$), deletion  and addition phases separate and a single direction dominates.

 \vspace{-1em}
\begin{figure}[!ht]
\begin{center}
    \begin{subfigure}{0.7\textwidth} 
        \begin{subfigure}{0.46\textwidth}
            \centering            \includegraphics[width=\linewidth]{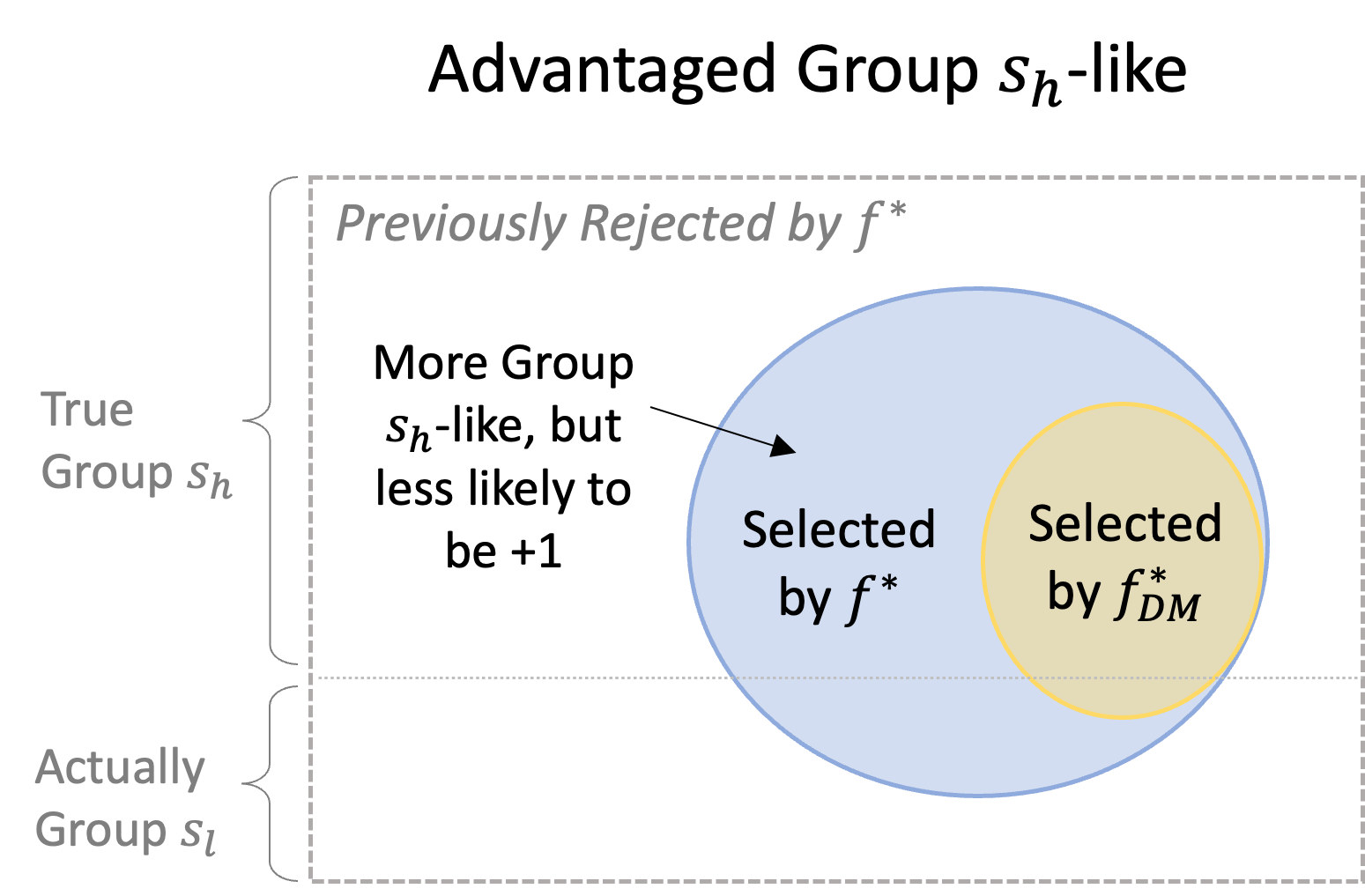}
            \caption{{$A_{\max} \leq B_{\min}$}}
            \label{fig:venn_blind_1}
        \end{subfigure}
        \hfill
        \begin{subfigure}{0.48\textwidth}
            \centering            \includegraphics[width=\linewidth]{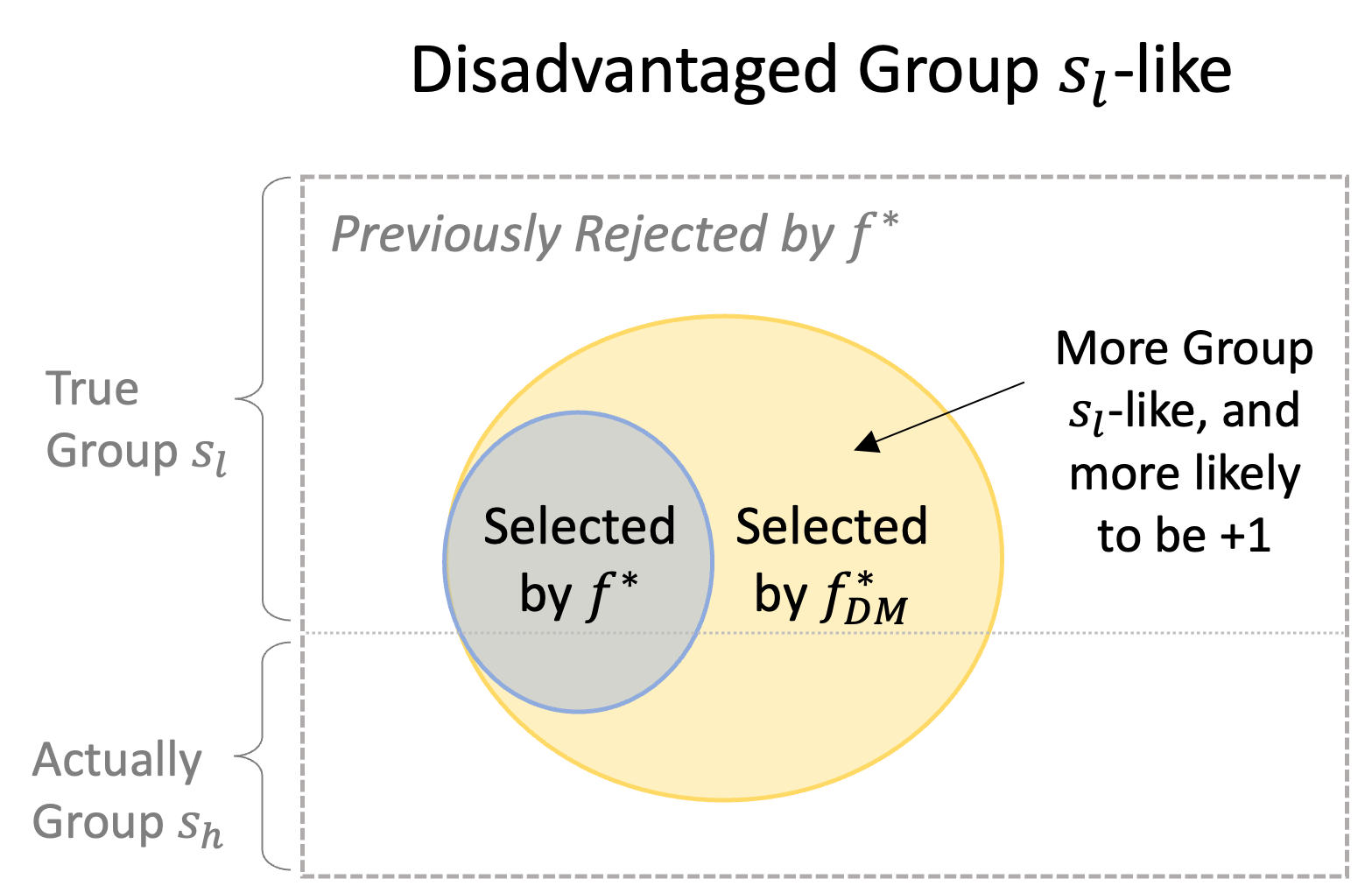}
            \caption{{$B_{\max} < A_{\min}$}}
            \label{fig:venn_blind_2}
        \end{subfigure}
    \end{subfigure}
    \caption{Selection Regions Before and After Enforcing Fairness in the Attribute-Blind Regime.} \label{fig:venn_blind}
    \end{center}
\end{figure}
\vspace{-1em}
\paragraph{Case $A_{\max}\le B_{\min}$ (deletion-only on the advantaged-like side).}
Under $A_{\max}\le B_{\min}$, the fair ML  \emph{reduces} the selected set \emph{exclusively} within the $\mathcal Q^{h}$  following the
order of $\zeta(x)$, which aggregates information of to what extent an $x$ are  advantaged-like and are marginal qualified under the unconstrained rule (how closed to the original boundary); see Figure \ref{fig:venn_blind_1}.
It leaves  other sides unaffected. 
Two consequences are immediate:
(i) Both groups' selection rates weakly decrease.  
The advantaged-like side contains members of both groups (true advantaged group members and masked candidates from the disadvantaged group).
Thus, deleting from this side can reduce  the selected mass of both groups 
(hence all group-wise selection rates, including NTR, weakly decrease).
(ii) Precision depends on how removals rank in $\eta$.
Fair ML removes previously selected candidates 
following the
order of $\zeta(x)$. 
Under order alignment conditions in Theorem \ref{thm:group_precision_noS}, one can compare the $\eta$ profiles of the removed portion to what remains, thus know how precision changes.

\paragraph{Case $B_{\max}\le A_{\min}$ (inclusion-only on the disadvantaged-like side).}
Under $B_{\max}\le A_{\min}$, the fair ML  \emph{expands} the selected set exclusively within $\mathcal Q^{l}$ by admitting previously unselected points following the order of $\zeta(x)$, 
see Figure \ref{fig:venn_blind_2}. 
Other sides remains unaffected. 
Two consequences are immediate: 
(i) Both groups' selection rates weakly increase. 
The disadvantaged-like side contains members of both groups; 
adding from this side can therefore  increase the selected mass of each group  (hence all group-wise selection rates, including NTR, weakly increase). 
(ii) Precision depends on how entrants rank in $\eta$. Fairness admits candidates following the order of $\zeta(x)$. 
Under conditions in Theorem~\ref{thm:group_precision_noS}, one can compare the $\eta$ profiles of the entrants to those of the incumbents; thus know how precision changes. 
\section{Conclusion}
In this paper, we adapt a unified population-level (Bayes) framework to assess how fair ML affects group outcomes in binary classification, covering both attribute-aware and attribute-blind deployment. In the attribute-aware regime, we show that enforcing common fairness notions necessarily (weakly) improves outcomes for the disadvantaged group while (weakly) reducing them for the advantaged group. In the attribute-blind regime, effects are distribution-dependent and may yield either leveling up or leveling down. 
We characterize when each pattern arises, highlight the role of ``masked'' candidates in driving them, 
and clarify the underlying mechanisms.

\vspace{6pt}
{\bibliographystyle{informs2014}
 
\bibliography{Ref_fairML}
}

\end{document}